\documentclass[times, review, 10pt]{elsarticle}

\usepackage[numbers]{natbib}

\def\tsc#1{\csdef{#1}{\textsc{\lowercase{#1}}\xspace}}
\tsc{WGM}
\tsc{QE}
\tsc{EP}
\tsc{PMS}
\tsc{BEC}
\tsc{DE}
\usepackage{textcomp}

\newcommand{\figref}[1]{Figure~\ref{#1}}%
\newcommand{\tableref}[1]{Table~\ref{#1}}%
\usepackage{multirow} 
\usepackage{tabularx}
\usepackage{threeparttable}
\usepackage{caption}
\usepackage{graphicx}

\usepackage{booktabs}
\usepackage{makecell}
\usepackage{tikz}
\usepackage{color}
\usepackage{colortbl}
\definecolor{lightgray}{gray}{0.85}
\definecolor{lightgreen}{RGB}{220, 245, 220}
\definecolor{lightorange}{RGB}{255, 220, 180}
\usepackage{amssymb}
\usepackage{pifont}
\usepackage{algorithm}
\usepackage{algorithmic}
\usepackage{subcaption}
 \usepackage{hyperref}
 \usepackage{amsmath}
 \usepackage{wasysym}
 \usepackage{xcolor}

\begin{document}

\begin{frontmatter}

\title{Unified Vision-Centric Pedestrian Crossing Action Prediction via Adaptive Patch Projection and Proactive Spatial Rectification}

\author[1]{Yao Tian}
\ead{2tianyao1@gmail.com}

\author[1]{Le Yang\corref{cor1}}
\ead{yangle@chd.edu.cn}

\author[2]{Binglu Wang}
\ead{wbl921129@gmail.com}

\cortext[cor1]{Corresponding author}

\address[1]{School of Electronics and Control Engineering, Chang'an University, Xi'an 710064, China}
\address[2]{School of Astronautics, Northwestern Polytechnical University, Xi'an 710072, China}

\begin{abstract}
Vision cues are available and informative for pedestrian action prediction, but obtaining stable target-centric representations from video frames remains challenging without frame-level external perception cues. Thus, most methods rely on additional perception modules or multi-source information fusion, leaving the reliability of vision-centric setting an open question.
To this end, we propose ViCross, a \textbf{vi}sion-centric pedestrian \textbf{cross}ing action prediction framework powered by multimodal large language models, which maintains target-centric reasoning from video frames without additional perception modules beyond first-frame target initialization.
While multimodal large language models exhibit strong visual understanding, applying them directly to vision-centric action prediction faces two challenges. First, accurately perceiving target pedestrians often requires high resolution inputs and dense visual tokenization, making full-frame encoding computationally prohibitive. ViCross tackles this with Variable Resolution Patch Mapping module for efficient token allocation while preserving key pedestrian details. Second, missing spatiotemporal priors hinder consistent cross frame reasoning. ViCross mitigates this with a Spatial Constraint Enhancement Strategy that captures past motion, future locations, and action semantics for training-time proactive spatial rectification.
Extensive experiments show that ViCross delivers clear gains in vision-centric prediction settings and is competitive with multi-source fusion approaches in several settings. Code is available at \href{https://github.com/2tianyao1/ViCross.git}{\textcolor{blue!65}{https://github.com/2tianyao1/ViCross.git}}.

\end{abstract}

\end{frontmatter}

\section{Introduction}
Pedestrian crossing action prediction is a critical component of safe autonomous driving systems. Reliable prediction requires understanding subtle cues related to pedestrian motion, body orientation, scene geometry, traffic conditions, and interactions with surrounding agents. 
Existing methods~\cite{chen2024pedestrian,kotseruba2021benchmark,yang2023dpcian} often address this problem through multi-source fusion pipelines that combine RGB frames with cues such as bounding boxes~\cite{elgazwy2025predicting}, pose~\cite{cadena2022pedestrian}, trajectory~\cite{rasouli2024diving,rasouli2023pedformer}, ego-vehicle speed, or semantic maps~\cite{sharma2025predicting}. Although these cues improve representation richness, they usually require additional perception modules or structured annotations during training and inference, which increases system complexity and may introduce error accumulation.

A more compact alternative is to formulate pedestrian crossing prediction as a vision-centric prediction problem, where the model predicts the crossing action mainly from video frames with minimal target initialization. However, this setting remains challenging. Without reliable frame-level target cues, the model must maintain attention on the same pedestrian across time while still capturing the surrounding context needed for action prediction.

Multimodal Large Language Models (MLLMs)~\cite{wang2024qwen2} provide a promising backbone for this setting because they can unify visual tokens and textual task instructions in a single reasoning framework. Nevertheless, directly applying MLLMs~\cite{huang2024gpt,ham2024omnipredict} to pedestrian crossing prediction introduces two key challenges. First, dense full-frame tokenization is computationally expensive, while coarse visual tokens may lose the fine pedestrian details required for target-centric reasoning. Second, when supervision is limited to crossing labels, the model may rely on scene or language priors rather than learning spatially grounded pedestrian motion.

\begin{figure}
\includegraphics[width=\textwidth]{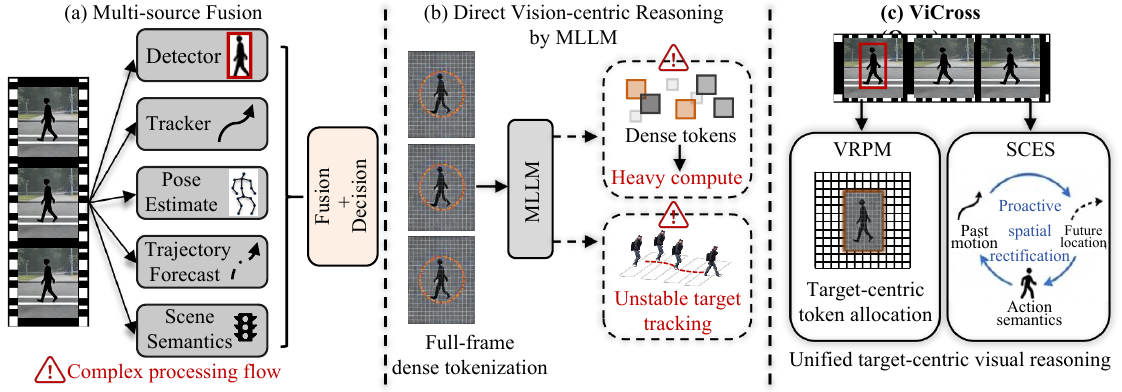}
\caption{Limitations of existing approaches and our solution.
Multi-source fusion pipelines involve complex processing flows and costly annotations, while direct full-frame MLLM reasoning incurs heavy computation and unstable target representations. ViCross enables efficient and stable target-centric reasoning without frame-level external perception modules.}
\label{intro}
\end{figure}

To address these issues, we propose ViCross, a unified vision-centric pedestrian crossing action prediction framework powered by MLLMs. As illustrated in \figref{intro}, ViCross differs from multi-source fusion pipelines and direct full-frame MLLM reasoning by reducing dependence on frame-level external perception modules while preserving target-centric pedestrian crossing prediction.
At the visual level, ViCross introduces a Variable Resolution Patch Mapping module that adaptively reorganizes each frame into a target-centric token grid. Regions around the target pedestrian and nearby traffic participants are encoded with finer patches, while distant background areas are represented more coarsely, preserving key appearance and motion cues with a controlled token budget.
In addition, ViCross incorporates a Spatial Constraint Enhancement Strategy that uses training-time auxiliary supervision. This provides proactive representation-level spatial rectification, encouraging shared visual tokens to align with pedestrian geometry and motion before the final crossing decision is generated.

Our contributions are threefold:

\begin{itemize} 
\item We introduce ViCross, a unified vision-centric framework for pedestrian crossing action prediction. With only first-frame pedestrian initialization, ViCross reduces dependence on frame-level external perception modules while retaining target-centric prediction.
\item We propose a Variable Resolution Patch Mapping module that preserves high spatial detail around the pedestrian while keeping the overall token budget compact and geometrically consistent.
\item We design a Spatial Constraint Enhancement Strategy that performs proactive representation-level spatial rectification through training-time supervision of past trajectory, future locations, crossing decisions, and text generation.
\end{itemize}

Experiments on JAAD\_beh, JAAD\_all, and PIE show that ViCross improves over vision-centric methods by 7\%, 6\%, and 2\%, respectively. Using only first-frame initialization, it remains competitive with several multi-source methods, reaching 74\% accuracy on JAAD\_beh and 90\% on JAAD\_all, while not matching the strongest multi-source methods on PIE.

\section{Related Work}

\subsection{Pedestrian Crossing Prediction}
Besides trajectory forecasting~\cite{zamboni2022pedestrian}, pedestrian crossing prediction is divided into intention prediction and action prediction~\cite{rasouli2024diving,gu2024orientation,wang2023temporal}. Intention prediction targets latent decision states that are difficult to define and annotate objectively, whereas action prediction focuses on observable crossing behaviors with clear labels. In practice, shared datasets and similar input–output formulations~\cite{rasouli2019pie,rasouli2017they} have blurred this distinction. In this work, we focus on pedestrian crossing action prediction, which aims to forecast whether a pedestrian will cross the street at a future time based on observations.

From the perspective of available information, mainstream pedestrian crossing prediction methods rely on multi-source information fusion~\cite{rasouli2021bifold,elgazwy2025predicting,liu2024camera}. Typically, RGB images~\cite{chen2024pedestrian,kotseruba2021benchmark} are combined with structured cues such as bounding boxes, segmentation maps, human pose, and ego-vehicle speed, using concatenation~\cite{singh2021multi}, multi-stream architectures~\cite{yang2022predicting}, or attention-based fusion~\cite{sharma2025predicting}. While these designs enrich scene representations, they incur substantial annotation and processing overhead, reducing efficiency and complicating practical deployment.
Some works instead restrict inputs to a single modality, such as static images~\cite{lorenzo2020rnn}, bounding boxes~\cite{achaji2022attention}, or pose cues~\cite{fang2018pedestrian}. Although simpler, these approaches struggle to model the spatiotemporal dynamics and agent-scene interactions underlying pedestrian crossing behavior, limiting robustness in real traffic scenarios.
Other methods attempt to reduce handcrafted multimodal inputs by learning from video frames~\cite{ahmed2023multi}, but in practice still rely on frame-level external perception modules, such as detectors~\cite{hou2025multispectral}, trackers~\cite{saleh2019real}, or pose estimators~\cite{neogi2020context}, including Faster R-CNN~\cite{ren2015faster}, YOLO~\cite{redmon2016you}, DeepSORT~\cite{wojke2017simple}, and OpenPose~\cite{cao2017realtime}.
In contrast, our model uses video frames with only first-frame target initialization, and performs target-centric reasoning without relying on external perception modules.

From a methodological perspective, early works~\cite{rasouli2019pie,lorenzo2020rnn} adopt CNNs combined with RNNs or LSTMs for spatiotemporal modeling, but are limited in capturing long-range dependencies and global context. Subsequent studies explore alternative temporal modeling paradigms, including 3D CNNs~\cite{singh2021multi,kotseruba2021benchmark}, recurrent variants such as GRUs~\cite{ham2023cipf}, and GCN-based approaches~\cite{chen2021visual,cadena2022pedestrian} that explicitly model inter-object relations but often introduce redundant interactions in crowded scenes.
Recently, Transformer architectures~\cite{zhang2023trep} have become dominant due to their effectiveness in modeling long-range temporal dependencies~\cite{zhou2023pit,elgazwy2025predicting}, with representative models such as PedFormer~\cite{rasouli2023pedformer} and IntentFormer~\cite{sharma2025predicting}. In parallel, MLLMs~\cite{wang2025multimodal,wang2024qwen2} have demonstrated strong global reasoning and multimodal representation capabilities.
Building on these advances, we introduce adaptive patch projection and a spatial constraint strategy to improve action reasoning and spatial grounding in MLLMs.


\subsection{Target-Centric Visual Representation Learning}

Target-centric representation learning biases visual encodings toward a specified object, rather than treating all regions uniformly. Existing methods fall into three categories. The first incorporates explicit spatial priors, such as RoI-based CNNs and object-conditioned queries in DETR~\cite{ren2015faster,meng2021conditional}, which focus computation on object regions but require accurate region supervision. The second uses structured attention to balance local and global context, including mechanisms that combine local-global interactions or hierarchical Transformers~\cite{wang2021pyramid,liu2021swin}, though they remain unaware of the target object. The third explores adaptive token strategies, such as learned sampling and dynamic sparsification~\cite{wang2023adaptive}, which highlight informative regions but often disrupt geometric consistency or rely on precise supervision, complicating integration with downstream temporal models. In this paper, we propose VRPM, which adapts patch resolution to the target region while maintaining a consistent panoramic token grid, enabling target-centric encoding without dense bounding-box supervision at inference.
\subsection{MLLMs for Vision-Centric Spatiotemporal Reasoning}
Multimodal Large Language Models (MLLMs)~\cite{wang2024qwen2} use large language models for unified sequence modeling, enabling visual and textual tokens to be processed within an auto-regressive framework. This approach supports temporally ordered visual signals without relying on task-specific architectures or handcrafted supervision. Recent advances have extended MLLMs from static images~\cite{liu2023visual} to videos~\cite{wang2025multimodal}, showing strong potential for modeling spatiotemporal patterns in RGB video sequences.
Conventional video Transformers or task-specific temporal classifiers~\cite{chen2024pedestrian,elgazwy2025predicting} typically map video features directly to a fixed binary prediction head. As a result, they often require specialized temporal designs or additional structured inputs~\cite{rasouli2021bifold} to capture target motion and scene interactions. In contrast, the MLLM conditions prediction on explicit task instructions and jointly processes visual tokens with textual prompts. This enables ViCross to identify the target pedestrian, track cross-frame motion, and reason over behavioral cues and scene context, without relying on structured inputs.


\section{Method}

\begin{figure}
\centering
\includegraphics[width
=\textwidth]{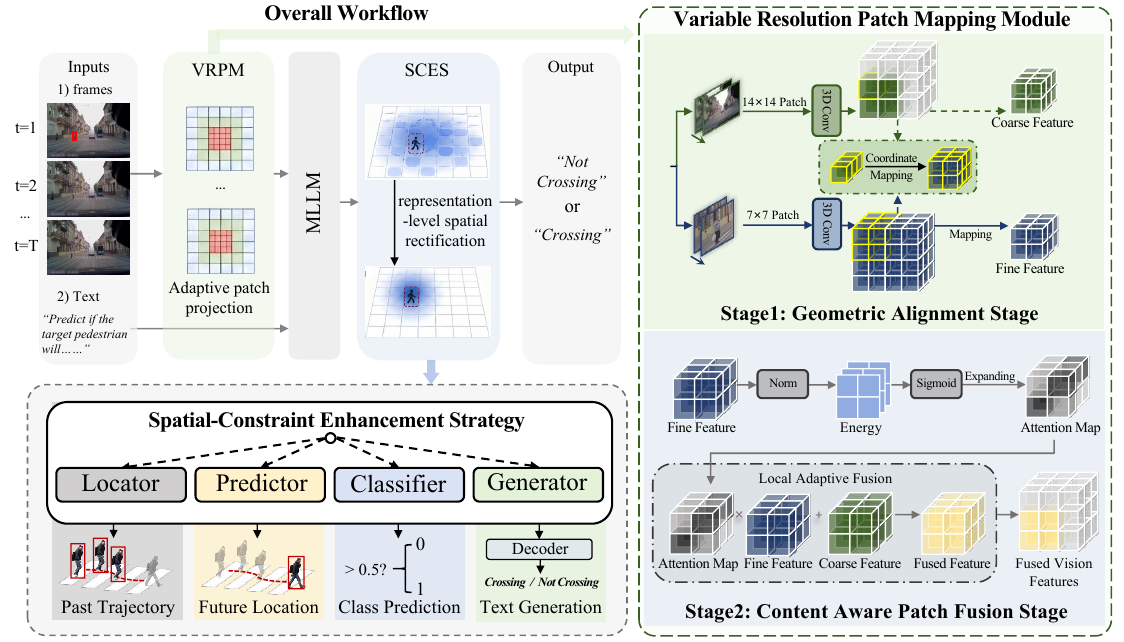}
\caption{Structure of ViCross. ViCross integrates two key components: Variable Resolution Patch Mapping and Spatial Constraint Enhancement Strategy. It takes a video and a fixed textual prompt as input, with only the target pedestrian annotated in the first frame, and predicts crossing behavior with auxiliary supervision from trajectory, future-position, and binary-decision tasks.}
\label{framework}
\end{figure}

\figref{framework} illustrates the overall architecture of ViCross. The model operates on video frames with the target pedestrian specified by a bounding box in the first frame.
ViCross consists of two components: (1) a Variable Resolution Patch Mapping module that aligns fine-grained pedestrian representations with global scene context while preserving spatial structure, and (2) a Spatial Constraint Enhancement Strategy that guides learning through joint prediction of past motion, future event location, and crossing action. 
Although bounding boxes provide auxiliary spatial supervision during training, ViCross only requires first-frame target initialization at inference time, avoiding frame-level detectors or trackers and simplifying the perception pipeline.
At a high level, VRPM and SCES serve complementary roles in ViCross. VRPM constructs target-centric visual tokens from the first-frame initialization by preserving fine pedestrian details and coarse scene context. These tokens are fed into the MLLM for reasoning, while SCES further regularizes them during training with spatial, motion, classification, and generation constraints. Notably, spatial rectification refers to representation-level regularization of shared visual tokens, not an inference-time localization correction. In this way, VRPM provides target-relevant visual evidence, SCES makes this evidence more spatially and temporally grounded, and the MLLM decoder generates the final crossing decision from the resulting representation.

\subsection{Problem Formulation}
Given an observed video clip $I=\left \{   i_{t-k+1},i_{t-k+2},...,i_t\right \}$ with $k$ frames, the task is to predict whether the target pedestrian will cross at future time $t+\tau$. The target is specified by an initial bounding box $B_0$ in the first frame. A fixed textual prompt $P$ describes the prediction objective. ViCross learns the mapping ${Z,\hat{L},\hat{F},A}=\mathcal{F}(I,B_0,P)$, where $Z=\left \{ z_1,z_2,...,z_n \right \} $ is decoded as the final textual decision. 

The auxiliary outputs include the observed trajectory $\hat{L}=\left \{  l^{t-k+1},...,l^t\right \} $, the future location $\hat{F}=l^{t+\tau}$, and the binary prediction $A\in\left \{   0,1\right \}$, which are used only for training supervision. 
\begin{algorithm}[htbp]
\caption{Variable Resolution Patch Mapping}
\label{alg:vrpm}
\begin{algorithmic}[1]
\REQUIRE Coarse features $F^{coarse} \in \mathbb{R}^{B \times T \times H_c \times W_c \times C}$, fine features $F^{fine} \in \mathbb{R}^{B \times T \times H_f \times W_f \times C}$, approximate pedestrian region $B_{0}^{pan} = (x_{0}^{pan} ,y_{0}^{pan} , x_{1}^{pan} , y_{1}^{pan} )$
\ENSURE Fused representation $F^{fused}$ aligned across resolutions

\STATE Initialize pooled tensor $F^{pooled} \leftarrow 0$ and mask $M \leftarrow 0$

\STATE \textbf{Stage 1: Geometric Alignment}
\FOR{each batch $b$ and frame $t$}
    \STATE Compute coarse grid range $(r_0,r_1,c_0,c_1)$ covered by $B_{0}^{pan}$
    \FOR{each coarse cell $(r,c)$ within this range}
        \STATE Map cell $(r,c)$ to fine grid coordinates in $\left [ f_{r,0}, f_{r,1} \right ] \times \left [ f_{c,0}, f_{c,1} \right ]$ via proportional scaling
        \STATE Select corresponding fine tokens $\Omega(r,c)$ within $\left [ f_{r,0}, f_{r,1} \right ] \times \left [ f_{c,0}, f_{c,1} \right ]$
        \STATE Aggregate fine tokens: $F^{pooled}_{b,t,r,c} \leftarrow \mathrm{mean}(F^{fine}_{b,t,\Omega(r,c)})$
        \STATE Mark $M_{b,t,r,c} \leftarrow 1$
    \ENDFOR
    \STATE For uncovered cells ($M_{b,t,r,c}=0$): fill $F^{pooled}_{b,t,r,c} \leftarrow F^{coarse}_{b,t,r,c}$
\ENDFOR

\STATE \textbf{Stage 2: Content Aware Fusion}
\STATE Compute attention map $A_{attn} = \sigma(\|F^{pooled}\|_2 / \sqrt{C}+\epsilon  )$
\STATE Expand $A_{attn}$ to match channel dimension
\STATE Fuse features at each spatial cell following the computation in \eqref{fused}.
\RETURN Fused representation $F^{fused}$
\end{algorithmic}
\end{algorithm}
The final crossing decision is obtained from the language-generation branch.
Compared with a direct binary classifier, the language-generation branch provides a prompt-conditioned decision pathway. The prompt specifies the target pedestrian and decision criterion, while autoregressive token supervision aligns the final output with task semantics. Thus, the auxiliary classifier serves as discriminative regularization, whereas the generator produces the final decision by using visual evidence, task instructions, and the spatiotemporal constraints.

\subsection{Variable Resolution Patch Mapping for Target-Centric Representation}
In pedestrian crossing scenarios, the target pedestrian occupies a small region of a high resolution frame and often appears near the visual field edges, making direct reasoning from global representations ambiguous. Inspired by human visual focusing~\cite{strasburger2011peripheral}, we propose the Variable Resolution Patch Mapping (VRPM) module, which assigns fine-grained patches to the inferred pedestrian region while encoding surrounding context with coarser patches. This design is useful when the target pedestrian is small, off-center, or surrounded by distracting backgrounds and nearby agents, where uniform full-frame tokenization either wastes computation on irrelevant areas or weakens target-centric visual reasoning. VRPM consists of two stages: geometric alignment and content aware patch fusion, as summarized in Algorithm~\ref{alg:vrpm}.

\subsubsection{Target Region Initialization}
To reduce frame-level annotation dependence, the target pedestrian is initialized by a first-frame box $B_{0}=\left(x_{0},y_{0},x_{1},y_{1}\right)$ and expanded by a shorter-side-proportional margin to tolerate temporal displacement:
\begin{equation}
\left\{
\begin{aligned}
w_{0} &= x_{1}-x_{0}, \quad
h_{0} = y_{1}-y_{0}, \quad
\delta = \gamma \min(w_{0},h_{0}), \\
x_{0}^{\gamma} &= x_{0}-\frac{\delta}{2}, \quad
y_{0}^{\gamma} = y_{0}-\frac{\delta}{2}, \\
x_{1}^{\gamma} &= x_{1}+\frac{\delta}{2}, \quad
y_{1}^{\gamma} = y_{1}+\frac{\delta}{2}.
\end{aligned}
\right.
\end{equation}
where $w_{0}$ and $h_{0}$ denote the width and height of $B_{0}$, and $\delta$ is the margin controlled by $\gamma$. The expanded region is then converted into a square box $B_{0}^{sq}=\left(x_{0}^{sq},y_{0}^{sq},x_{1}^{sq},y_{1}^{sq}\right)$:
\begin{equation}
\begin{cases}
  & w_{0}^{\gamma} =x_{1} ^{\gamma}-x_{0} ^{\gamma},  h_{0}^{\gamma}=y_{1} ^{\gamma}-y_{0} ^{\gamma}  \\
  & x_{0}^{sq} =x_{0} ^{\gamma}- \frac{h_{0}^{\gamma}- w_{0}^{\gamma} }{2},y_{0}^{sq} =y_{0} ^{\gamma}  \\
  & x_{1}^{sq} =x_{1} ^{\gamma}+ \frac{h_{0}^{\gamma}- w_{0}^{\gamma} }{2},y_{1}^{sq} =y_{1} ^{\gamma}
\end{cases}
\end{equation}
where $w_{0}^{\gamma}$ and $h_{0}^{\gamma}$ denote the width and height of $B_{0}^{\gamma}$ in raw image coordinates. 
Based on 
$B_{0}^{sq}$, we clip its coordinates to the image boundary as $B_{0}^{c}=\left ( x_{0}^{c},y_{0}^{c},x_{1}^{c},y_{1}^{c}  \right )$:
\begin{equation}
\begin{cases}
  & x_{0}^{c}=\max(0,x_{0}^{sq}), y_{0}^{c}=\max(0,y_{0}^{sq}) \\
  & x_{1}^{c}=\min(W,x_{1}^{sq}), y_{1}^{c}=\min(H,y_{1}^{sq})
\end{cases}
\end{equation}
where $W=1920$ and $H=1080$ denote the raw image width and height. 

The enlarged square region serves as a lightweight target-centric spatial prior. Within the short 16-frame observation window, it provides an approximate local anchor for moderate image-plane displacement and scale variation. This design increases the spatial margin around the initially tall and narrow pedestrian box. More complex dynamic localization modules may provide more accurate target coverage, but would introduce additional computation and perception dependency. Taking this trade-off into account, we adopt the current design as a practical compromise, and its effectiveness is demonstrated by the experimental results.
Nevertheless, its robustness still depends on first-frame initialization quality. Severe shifts, heavy occlusion, or overlap with nearby pedestrians may introduce incomplete or misleading target information into the local high-resolution crop, reducing the reliability of pedestrian-specific tokens and weakening target-centric reasoning in subsequent frames.

\subsubsection{Geometric Alignment Stage}
\begin{figure}
\centering
\includegraphics[width=\textwidth]{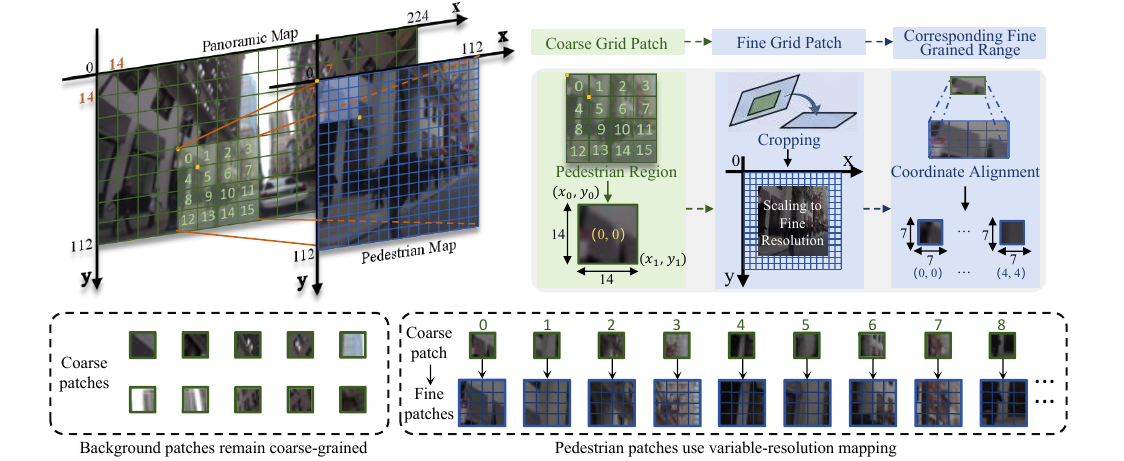}
\caption{Detailed operation of the coordinate mapping. The panoramic map and the approximate pedestrian-local map are partitioned into patches at different scales. Coarse-grained patches within the pedestrian region are mapped to their corresponding fine-grained regions via coordinate transformation.}
\label{VRPM}
\end{figure}

As shown in \figref{VRPM}, VRPM aligns fine-grained pedestrian patches onto a coarse panoramic grid, while keeping background regions unchanged in the coarse grid. Rather than using direct multi-scale fusion such as concatenation or cross attention, it establishes explicit spatial correspondence between coarse patches and their covered fine-grained regions. The aligned fine features are aggregated onto the coarse grid, yielding a unified two-dimensional patch layout.

First, the panoramic input $I_{pan}$ is partitioned using a coarse patch size $P_c$. Based on the pedestrian region $B_{0}^{c}$, the coarse patches covering the pedestrian are identified, and their boundaries are used to adjust $B_{0}^{c}$, yielding the refined pedestrian region $B_{0}^{pan}=(x_{0}^{pan} ,y_{0}^{pan} , x_{1}^{pan} , y_{1}^{pan} )$. The high-resolution pedestrian image $I_{ped}$ is then cropped from the input frame based on $B_{0}^{pan}$.

Second, $I_{ped}$ is partitioned using a finer patch size $P_{f} < P_{c}$. $I_{pan}$ and $I_{ped}$ are encoded by 3D convolutions to obtain coarse-grained and fine-grained feature maps, $F^{coarse} \in \mathbb{R}^{B \times T \times H_c \times W_c \times C}$ and $F^{fine} \in \mathbb{R}^{B \times T \times H_f \times W_f \times C}$. Here, $T$ is the number of frames, $H\times W$ the spatial grid size, and $C$ the feature dimension.
For each covered coarse patch $\left ( r,c \right ) $, its corresponding pixel range $\left [ bx_{0},bx_{1} \right )\times \left [ by_{0},by_{1} \right )$ on $I_{pan}$ is computed.
We project pixel range $\left [ bx_{0},bx_{1} \right )\times \left [ by_{0},by_{1} \right )$ onto the normalized cropped image space, assuming a size of $112\times 112$, to obtain the relative coordinates $\left [ rel\_x_{0},rel\_x_{1} \right )\times \left [ rel\_y_{0},rel\_y_{1} \right )$:
\begin{equation}
\begin{cases}
  & rel\_x_{k} =\frac{bx_{k}-x_{0}^{pan}  }{x_{1}^{pan}-x_{0}^{pan}} \times 112 ,  \,k\in\left \{ 0,1 \right \}  \\
  & rel\_y_{k} =\frac{by_{k}-y_{0}^{pan}  }{y_{1}^{pan}-y_{0}^{pan}} \times 112, \,k\in\left \{ 0,1 \right \},
\end{cases}
\end{equation}
Third, fine-grained patches are identified and aggregated via average pooling. The coordinates $\left [ rel\_x_{0},rel\_x_{1} \right )\times \left [ rel\_y_{0},rel\_y_{1} \right )$ are mapped to the fine-grained patch index range $\left [ f_{r,0}, f_{r,1} \right ] \times \left [ f_{c,0}, f_{c,1} \right ]$ according to the fine-grained patch size $P_{f}$.

Finally, the aligned feature $F_{pooled}$ is computed by average pooling the fine-grained features $F^{fine} \in \mathbb{R}^{B \times T \times H_f \times W_f \times C}$ corresponding to $\left [ f_{r,0}, f_{r,1} \right ] \times \left [ f_{c,0}, f_{c,1} \right ]$:
\begin{equation}
F^{pooled} \left [ b,t,r,c,: \right ] =\frac{1}{\left | S \right | } \sum_{\left ( r^{'},c^{'} \right )\in S }^{ }F^{fine}\left [ b,t,r^{'},c^{'},:  \right ],
\end{equation}
where $S=\left [ f_{r,0}, f_{r,1} \right ] \times \left [ f_{c,0}, f_{c,1} \right ]$. A binary coverage mask $M\in \left \{ 0,1 \right \} ^{B\times T\times H_{c}\times W_{c} } $ is maintained alongside, with $M_{b,t,r,c} =1$ denotes a covered coarse patch.

\subsubsection{Content Aware Patch Fusion Stage}
Following geometric alignment, content aware patch fusion adaptively weights fine-level and coarse-level patches to emphasize discriminative pedestrian cues and suppress background interference.

After flattening $F^{coarse}$, $M$, and $F^{pooled}$ to achieve the shape $\mathbb{R}^{B \times T \times C \times H_{p} \times W_{p}}$, fusion is then performed at the coarse-grained graph scale. The attention weight $A_{attn}$ is calculated based on the feature vector norm of the aligned fine-grained feature $F^{pooled}$ to perceive its information intensity:
\begin{equation}
A_{attn}=\sigma \left ( \frac{\left \| F^{pooled}  \right \|_{2}  }{\sqrt{C}+\epsilon  }  \right ), 
\end{equation}
where $\left \| \cdot  \right \| _{2} $ is the L2 norm, and $\sigma\left ( \cdot  \right ) $ is the Sigmoid activation function. The final fused feature $F^{fused}\in \mathbb{R}^{B\times T\times C\times H_{p}\times W_{p}  } $ undergoes attention-weighted summation in the pedestrian region (marked by mask $M$), and the coarse-grained feature $F^{coarse}$ is kept unchanged in the non-pedestrian region:
\begin{equation}
F^{fused} = 
\begin{cases}
A_{attn} \odot F^{pooled} + F^{coarse}, & \text{if } M=1 \\
F^{coarse}, & \text{otherwise},
\end{cases}
\label{fused}
\end{equation}

So far, we obtain the fused representation $F^{fused}$ that integrates coarse- and fine-grained information. VRPM enforces cross-scale geometric consistency while adaptively emphasizing discriminative regions.

\begin{figure}
\centering
\includegraphics[width=\textwidth]{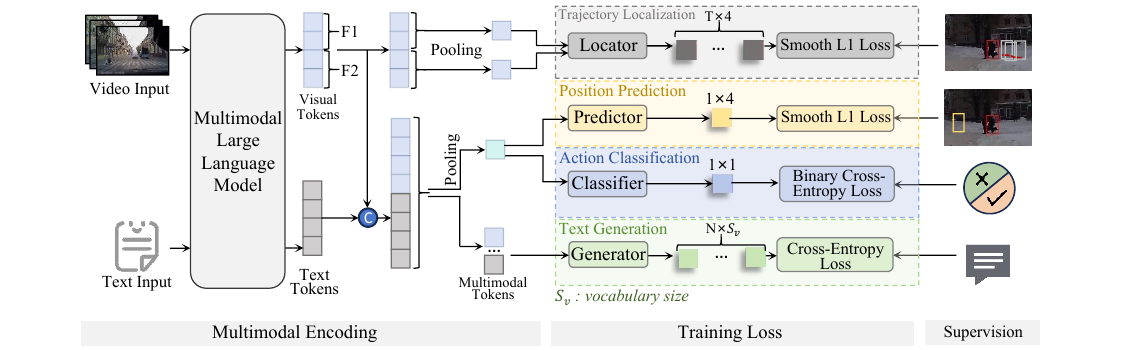}
\caption{Detailed design of the Spatial Constraint Enhancement Strategy. The locator, predictor, classifier, and generator provide training-time spatial rectification by enforcing target localization, future-motion regularization, crossing-state discrimination, and language-based alignment. These constraints improve target-centric and motion-aware encoding without requiring dense spatial annotations during inference.}
\label{sces}
\end{figure}

\subsection{Spatial Constraint Enhancement Strategy}
Crossing prediction relies on stable spatiotemporal representations of the target pedestrian across frames. In a unified MLLM-based framework without frame-level external perception inputs, explicit spatial and motion priors are no longer enforced, often resulting in weak cross-frame consistency and unstable target focus in dynamic scenes.
To address this issue, we introduce a Spatial Constraint Enhancement Strategy (SCES) to perform proactive representation-level spatial rectification during training. Different from a standard multi-task design that only adds auxiliary predictions, SCES constrains the shared visual tokens used by the generator to recover observed pedestrian locations, anticipate future positions, and distinguish crossing states. These constraints rectify the latent representation before the final language-based decision is generated, making the prediction spatially grounded rather than driven only by binary supervision.

SCES introduces spatial supervision through joint training objectives. Specifically, the overall objective is formulated as a weighted sum of individual loss components:
\begin{equation}
\mathcal{L}_{Total} =\omega _{1}\mathcal{L}_{LM}+\omega _{2}\mathcal{L}_{CLS}+\omega _{3}\mathcal{L}_{PRE}+\omega _{4}\mathcal{L}_{OBS}, 
\end{equation}
where $\mathcal{L}_{OBS}$ and $\mathcal{L}_{PRE}$ denote the bounding box losses for past observations and future trajectory predictions, respectively, $\mathcal{L}_{CLS}$ is the binary loss, and $\mathcal{L}_{LM}$ is the language modeling loss, with non-negative weights balancing the four objectives.

\textbf{Text generation.}
The generator produces the final textual crossing decision through the LLM decoder. This branch follows the full prompt-conditioned MLLM decoding path, so the final prediction is jointly conditioned on visual evidence, task instructions, and semantic constraints. The base loss is the standard Cross-Entropy token prediction loss applied to the text tokens:
\begin{equation}
\mathcal{L}_{LM} = - \frac{1}{N_{LM} } \sum_{i=1}^{N_{LM} } \sum_{v\in \mathcal{V}}\mathbb{I}(y_i = v)log\left ( P\left ( v\mid x_{< i}  \right )  \right ) ,
\end{equation}
where $N_{LM}$ is the number of valid target tokens, $\mathcal{V}$ is the vocabulary, $\mathbb{I}\left ( \cdot \right )$ is the indicator function, and $P\left ( v\mid x_{< i} \right )$ denotes the predicted probability of token $v$ given the preceding context $x_{< i}$. In language modeling, $x_{< i}$ refers to all tokens preceding position $i$. The model adopts an autoregressive next-token prediction formulation, where the output at position $i$ is used to predict the token at position $i+1$.

\textbf{Crossing action classification.}
The classifier predicts a binary crossing action from aggregated representations. Since it is a lightweight binary head rather than the full MLLM decoding path, directly using it as the final output would prevent the decision from fully leveraging prompt semantics and language constraints in the generative reasoning process. Therefore, this branch provides discriminative supervision during training but is not used as the final decision head. In addition, the model predicts the logit $z$ for a binary classification task. The loss $\mathcal{L}_{CLS}$ uses the $\textit{BCEWithLogitsLoss}$ for numerical stability, which combines the sigmoid function $\sigma \left ( \cdot  \right ) $ and Binary Cross-Entropy. A positive weight $\varepsilon  $ is incorporated to address potential class imbalance, corresponding to the positive class weight. The formula is as follows:
\begin{equation}
\mathcal{L}_{CLS} =  - y \cdot \varepsilon  log\left ( \sigma \left ( z  \right )  \right) -\left ( 1-y  \right )\cdot log\left ( 1-\sigma \left ( z  \right )  \right ) ,
\end{equation}
where $y$ is the ground truth and $\sigma \left ( \cdot  \right ) $ is the sigmoid function.

\textbf{Future position prediction.}
The predictor estimates the pedestrian position at a future time from globally aggregated representations, regularizing future motion dynamics and encouraging motion-aware visual representations. This objective is used to contribute to proactive representation-level spatial rectification by regularizing future motion dynamics during training. During inference, the predicted future position can be used for visualization and interpretation, but it is not used as the final crossing decision or fed back into the generator.
Bounding box regression task employs the Smooth L1 Loss, commonly known as Huber Loss, defined as $Huber(p,t)$ with the parameter $\beta $ set to 1.0. The Huber loss function for a residual $r= \left ( p-t \right ) $ is:
\begin{equation}
Huber\left ( p,t \right ) =\begin{cases}
  0.5\left ( p-t \right )^{2}  & \text{ if } \left | p-t \right |< \beta  \\
  \beta \left| p-t \right |-0.5\beta ^{2}  & otherwise,
\end{cases}
\end{equation}
The loss $\mathcal{L}_{PRE}$ computes the regression error between the predicted final crossing point bounding box $P$ and its ground truth counterpart $G$. The box coordinates are typically represented by 4 values $\left [ x_{min}, y_{min},x_{max}, y_{max} \right ] $.
\begin{equation}
\mathcal{L}_{PRE}=\frac{1}{4} \sum_{k=1}^{4} Huber\left ( P_{k},G_{k}  \right ) ,
\end{equation}
where $P_{k}$ and $G_{k}$ are the k-th predicted and ground truth coordinates, respectively.

\textbf{Past trajectory localization.}
The locator regresses frame-level pedestrian bounding boxes from visual tokens, enforcing target-centric spatial focus over the observed frames and improving cross-frame consistency. Similarly, the loss $\mathcal{L}_{OBS}$ calculates the error between the bounding boxes predicted from the $T$ observed visual tokens and their corresponding ground truth bounding boxes $G^{'}$:
\begin{equation}
\mathcal{L}_{OBS}=\frac{1}{4\cdot T} \sum_{t=1}^{T}\sum_{k=1}^{4} Huber\left ( P_{t,k},G_{t,k}  \right ) ,
\end{equation}
where $P_{t,k}$ and $G_{t,k}$ are the $k$-th predicted and ground truth coordinates at observation step $t$. The internal design details of SCES are shown in \figref{sces}.

During inference, SCES requires no frame-level bounding boxes or auxiliary spatial labels. The locator, predictor, and classifier are used only for training-time spatial constraints, not for explicit localization correction at test time. Their effects are encoded in the shared visual representations, from which the language decoder generates the final crossing decision using visual tokens and the textual prompt. Since localization errors are not directly corrected during inference, target drift and spatial misalignment may still occur under inaccurate initialization or severe occlusion.

\section{Experiments}
\subsection{Datasets}
JAAD~\cite{rasouli2017they} and PIE~\cite{rasouli2019pie} are public datasets designed for pedestrian crossing action prediction in traffic scenarios. The JAAD dataset includes 323 valid videos after excluding clips captured at low resolution or under unfavorable conditions. It is organized into two subsets: JAAD\_beh, which contains pedestrians that are crossing or about to cross, and JAAD\_all, which augments JAAD\_beh with 2,100 non-crossing pedestrians. JAAD is collected under diverse weather and scene complexities, thereby faithfully representing real-world driving conditions. 
In addition, the PIE dataset consists of six hours of driving footage recorded under ideal weather conditions, featuring a larger number of annotated pedestrians, longer tracking sequences, and ego-vehicle speed data obtained via on-board diagnostics  sensors.
Following Kotseruba et al.~\cite{kotseruba2021benchmark}, we adopt train/test splits of 194/171, 783/612, and 795/636 for JAAD\_beh, JAAD\_all, and PIE, respectively. Samples are generated with overlap ratios of 0.8, 0.8, and 0.6, while labels are assigned according to the official JAAD and PIE protocols~\cite{rasouli2017they,rasouli2019pie}.

\begin{table*}[htbp]
  \centering
  \caption{Comparison with existing pedestrian crossing prediction methods. We report each method’s architecture, direct model inputs, and 2D bounding-box usage. ``Input Types'' denotes explicit model inputs, while ``2D BBOX'' denotes whether the input frames are cropped, initialized, or otherwise processed using 2D bounding boxes. We report accuracy, F1, precision, and recall.}
  \renewcommand{\arraystretch}{1.5} 
  \setlength{\tabcolsep}{4pt} 
  \begin{threeparttable}   
  \resizebox{\textwidth}{!}{%
    \begin{tabular}{c|c|c|c|rrrr|rrrr|rrrr}
    \hline
    \multirow{2}[4]{*}{Method} & \multirow{2}[4]{*}{Model Variants} & \multirow{2}[4]{*}{Input Types}& \multirow{2}[4]{*}{\shortstack{2D\\ BBOX}} & \multicolumn{4}{c|}{JAAD\_beh} & \multicolumn{4}{c|}{JAAD\_all} & \multicolumn{4}{c}{PIE} \\
\cline{5-16}          &       &       &       & Acc & F1 & Prec & Rec & Acc & F1 & Prec & Rec & Acc & F1 & Prec & Rec \\
    \hline
    \multicolumn{15}{l}{\textbf{Multi-source Fusion Methods}} \\
    \hline
    MultiRNN~\cite{bhattacharyya2018long} & GRU   & I,2DB,EVS &   \scalebox{1.3}{$\circ$}    &    0.61    &    0.74    &    0.64    &    0.86    &    0.79    &    0.58    &    0.45    &    0.79    &    0.83    &    0.71    &    0.69    &    0.73 \\
    PCPA~\cite{kotseruba2021benchmark}  & 3D Conv & I,PO,2DB,EVS &   \scalebox{1.4}{$\bullet$}    &  0.50  &  0.59  &  0.61  &  0.58  &  0.70  &  0.51  &  0.36  &  0.87  &  0.86  &  0.78  &  0.69  &  0.89 \\
    FFSTA~\cite{yang2022predicting} & Self Attention & I,PO,2DB,EVS &   \scalebox{1.4}{$\bullet$}    &  0.62  &  0.74  &  0.65  &  0.85  &  0.83  &  0.63  &  0.51  &  0.81  &  0.89  &  0.80  &  0.79  &  0.81 \\
    V-PedCross~\cite{bai2022deep} & 3D Conv & I,2DB,Flo &   \scalebox{1.4}{$\bullet$}    &  0.64  &  0.76  &  0.70  &  0.89  &  0.86  &  0.77  &  0.74  &  0.81  &  0.89  &  0.67  &  0.74  &  0.84 \\
    Ped Graph+~\cite{cadena2022pedestrian} & GCN   & I,Seg,PO,EVS &   \scalebox{1.4}{$\bullet$}    &  0.70  &  0.76  &  0.77  &  0.75  &  0.86  &  0.65  &  0.58  &  0.75  &  0.89  &  0.81  &  0.83  &  0.79 \\
    PedGNN~\cite{riaz2023synthetic} & GRU   & PO  &   -    &   -   &   -   &   -   &   -   &   0.86   &   0.86   &   0.77   &   0.96   &   0.71   &   0.79   &   0.75   &   0.83 \\
    Tamformer~\cite{osman2023tamformer} & Transformer & I,2DB,PO &   \scalebox{1.4}{$\bullet$}    &   0.73   &   0.80   &   -   &   -   &   0.88   &   0.68   &   -   &   -   &   0.88   &   0.79   &   -   &   - \\
    DPCIAN~\cite{yang2023dpcian} & GCN   & I,PO  &   \scalebox{1.3}{$\circ$}     &   0.71   &   0.78   &   0.73   &   0.85   &   0.89   &   0.59   &   0.61   &   0.58   &   0.91   &   0.83   &   0.83   &   0.84 \\
    PIT~\cite{zhou2023pit}   & Transformer & I,PO,EVS,2DB &   \scalebox{1.4}{$\bullet$}    &   0.70   &   0.81   &   0.71   &   0.93   &   0.87   &   0.66   &   0.54   &   0.85   &   0.91   &   0.82   &   0.85   &   0.79 \\
    TREP~\cite{zhang2023trep}  & Transformer & 2DB,EVS &   -    &   -   &   -   &   -   &   -   &   0.88   &   0.61   &   0.70   &   0.54   &   0.92   &   0.85   &   0.88   &   0.82 \\
    PedAST-GCN~\cite{ling2024pedast} & STA-GCN & 2DB,PO,EVS &   -    &   0.69   &   0.79   &   0.68   &   0.93   &   0.89   &   0.68   &   0.67   &   0.69   &   0.91   &   0.83   &   0.88   &   0.79 \\
    Faster-PCPnet~\cite{yang2024faster} & Conv+GCN & 2DB,PO,EVS &   -    &  -  &  -  &  -  &  -  &  0.89  &  0.65  &  0.73  &  0.58  &  0.94  &  0.89  &  0.89  &  0.88 \\
    \hline
    \multicolumn{15}{l}{\textbf{Vision-Centric Methods}} \\
    \hline
    \multirow{2}{*}{\centering Static}& VGG-16~\cite{simonyan2015very} & I     &   \scalebox{1.4}{$\bullet$}    &  0.59 &  0.71 &  0.63 &  0.82 &  0.82 &  0.55 &  0.49 &  0.63 &  0.71 &  0.41 &  0.49 &  0.36 \\
         & ResNet50~\cite{he2016deep} & I     &   \scalebox{1.4}{$\bullet$}    &  0.46 &  0.54 &  0.58 &  0.51 &  0.81 &  0.52 &  0.47 &  0.56 &  0.70 &  0.38 &  0.47 &  0.32 \\
    \hline
    \multirow{2}{*}{\centering ConLSTM~\cite{shi2015convolutional}}  & VGG16+LSTM & I     &   \scalebox{1.4}{$\bullet$}    &  0.53  &  0.64  &  0.64  &  0.64  &  0.63  &  0.32  &  0.24  &  0.48  &  0.58  &  0.39  &  0.32  &  0.49 \\
& ResNet50+LSTM & I     &   \scalebox{1.4}{$\bullet$}    &  0.59 &  0.69 &  0.68 &  0.70 &  0.63 &  0.33 &  0.25 &  0.49 &  0.54 &  0.26 &  0.23 &  0.29 \\
    \hline
    I3D~\cite{carreira2017quo}   & 3D Conv & I     &   \scalebox{1.4}{$\bullet$}     &    0.62 &    0.73 &    0.68 &    0.79 &    0.81 &    0.63 &    0.66 &    0.61 &    0.80 &    0.62 &    0.67 &    0.58 \\
    C3D~\cite{tran2015learning}   & 3D Conv & I     &   \scalebox{1.4}{$\bullet$}     &  0.61 &  0.75 &  0.63 &  0.91 &  0.84 &  0.65 &  0.57 &  \textbf{0.75} &  0.77 &  0.52 &  0.63 &  0.44 \\
    Intentformer~\cite{sharma2025predicting} & Transformer & I     &   \scalebox{1.4}{$\bullet$}    &  0.45  &  -  &  -  &  -  &  0.60  &  -  &  -  &  -  &  0.59  &  -  &  -  &  - \\
    GPT4V TW~\cite{huang2024gpt} & GPT-4V   & I     &   \scalebox{1.4}{$\bullet$}    &  0.57 &  0.65 &  \textbf{0.82} &  0.54 &  - &  - &  - &  - &  - &  - &  - &  - \\
    GPT4Vskip TW~\cite{huang2024gpt} & GPT-4V   & I     &   \scalebox{1.4}{$\bullet$}    &  0.55 &  0.64 &  0.81 &  0.53 &  - &  - &  - &  - &  - &  - &  - &  - \\
    Omnipredict~\cite{ham2024omnipredict} & GPT-4o   & I     &   \scalebox{1.4}{$\bullet$}    &  0.67 &  0.65 &  0.66 &  0.65 &  - &  - &  - &  - &  - &  - &  - &  - \\
    \hline
    \multicolumn{1}{c|}{\cellcolor{lightgray}ViCross}   & {\cellcolor{lightgray}MLLM}   & {\cellcolor{lightgray}I}     &   {\cellcolor{lightgray}{\fontsize{9pt}{9pt}\selectfont $\LEFTcircle$} }    &  {\cellcolor{lightgray} \textbf{0.74}}     & {\cellcolor{lightgray} \textbf{0.82}}      & {\cellcolor{lightgray} 0.72}      &  {\cellcolor{lightgray} \textbf{0.97}}     & {\cellcolor{lightgray} \textbf{0.90}}      & {\cellcolor{lightgray} \textbf{0.71}}      &   {\cellcolor{lightgray} \textbf{0.73}}    &  {\cellcolor{lightgray} 0.69}     &{\cellcolor{lightgray} \textbf{0.82}}       &  {\cellcolor{lightgray} \textbf{0.66}}     & {\cellcolor{lightgray} \textbf{0.69}}      &{\cellcolor{lightgray} \textbf{0.64}}  \\
    \hline
    \end{tabular}%
    }
    \vspace{3pt}
\begin{tablenotes}[flushleft]
  \footnotesize
  \item \parbox{\textwidth}{
    \textbf{Input Types}: RGB image (\textbf{I}); 2D bounding box (\textbf{2DB});
    Pose (\textbf{PO}); Ego-vehicle speed (\textbf{EVS}); Optical flow (\textbf{Flo});
    Segmentation map (\textbf{Seg}).
2D BBOX: whether the input video frames are annotated or processed with 2D bounding boxes. \scalebox{1.4}{$\bullet$}: 2D bounding boxes are applied to all input frames; {\fontsize{5pt}{5pt}\selectfont $\LEFTcircle$}: a 2D bounding box is applied only to the first input frame for target initialization; \scalebox{1.3}{$\circ$}: no 2D bounding box is applied to the input frames.
  }
\end{tablenotes}

  \end{threeparttable}
  \label{comparision_table}%
\end{table*}%

\subsection{Implementation Details}

The loss weights $(\omega_1,\omega_2,\omega_3,\omega_4)$ are set to $(1,0.5,0.5,0.5)$ for JAAD\_beh and JAAD\_all, and $(1,1,1,1)$ for PIE. Since $\mathcal{L}_{LM}$ corresponds to the final generation-based prediction, its weight is fixed to 1. 
The auxiliary losses use shared weights to balance spatial and semantic supervision across branches, with weights selected via constrained search. We adopt Qwen2-VL-7B as the MLLM backbone. For efficient adaptation, we train only LoRA adapters in the visual backbone and LLM, the 3D convolution layers in VRPM, as well as the locator, predictor, and classifier in SCES.
For training, we use AdamW with a learning rate of $1\times10^{-4}$, batch size of 16, 10 epochs, and random seed 42 on all datasets. Main tables use seed 42, while twelve-seed runs are reported only for stability analysis.
Experiments are conducted on customized NVIDIA RTX 4090 GPUs with approximately 48 GiB memory. During inference, the panoramic input is set to $448\times224$ with patch size 14, and the pedestrian input to $112\times112$ with patch size 7. Results are reported using the final-stage checkpoint.

\textbf{Inference.} Given a test video, only the first-frame target box is used for initialization. Following Sections 3.2.1 and 3.2.2, we construct the pedestrian-centered region and apply VRPM to generate panoramic and pedestrian tokens. The fused visual tokens, together with the fixed textual prompt, are fed into the MLLM for prediction. The final crossing prediction is obtained from the generated textual response. The fixed prompt requires the model to output exactly ``crossing'' or ``not crossing''. During evaluation, only exact matches are accepted and mapped to the corresponding binary labels. Any other output is treated as invalid and assigned the opposite ground-truth label, thus counted as incorrect. No invalid outputs are observed after training.


\begin{figure}
\centering
\includegraphics[width=\textwidth]{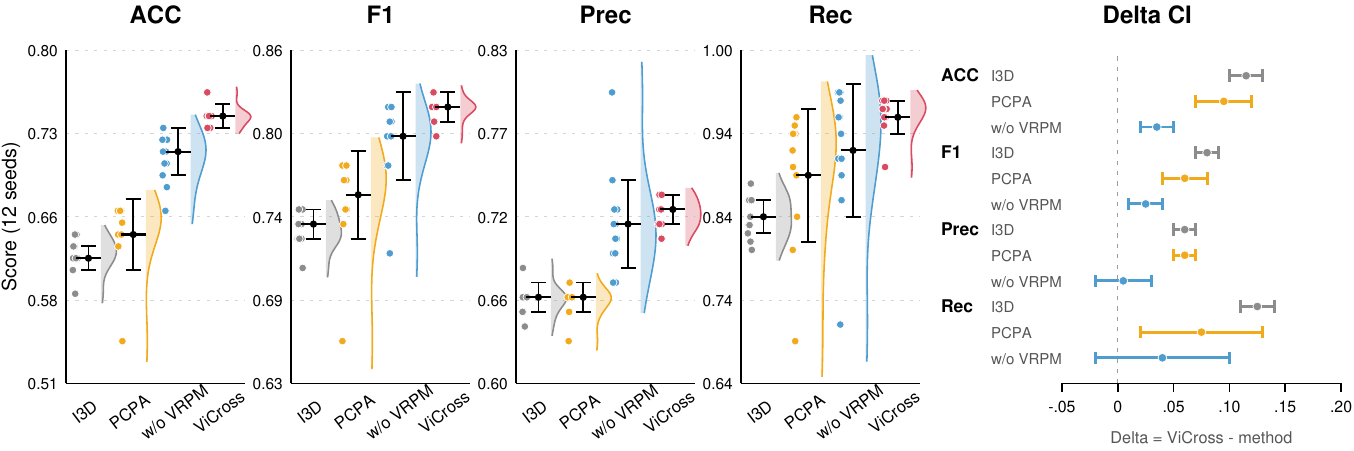}
\caption{
Stability with 95\% CI across twelve seeds. Colored points denote individual seed results, half-violins show Gaussian kernel density estimates, and black markers show mean $\pm$ standard deviation. The right panel reports Delta with 95\% CI, measuring ViCross improvement over compared method.
}
\label{cloudrain}
\end{figure}

\subsection{Comparison with Vision-Centric and Multi-source Prediction Methods}

Following the previous protocol~\cite{kotseruba2021benchmark,yang2024faster}, each sample contains 16 observation frames with a TTE of 30 to 60 frames. We report accuracy, F1, precision, and recall, but omit AUC because the final decision is derived from generated text rather than calibrated class probabilities. Baseline results are taken from the corresponding papers, with Static, ConLSTM, C3D, and I3D adopted from Kotseruba et al.~\cite{kotseruba2021benchmark}.
\tableref{comparision_table} compares ViCross with vision-centric and multi-source fusion methods. Vision-centric methods mainly use RGB frames, while multi-source methods often rely on frame-level boxes, pose, ego-vehicle speed, segmentation, or optical flow. Under the vision-centric setting, ViCross consistently improves over baselines across all three datasets, achieving 0.74, 0.90, and 0.82 Acc on JAAD\_beh, JAAD\_all, and PIE, respectively.

Compared with multi-source methods~\cite{kotseruba2021benchmark,cadena2022pedestrian,bhattacharyya2018long}, ViCross follows a simpler vision-centric input setting. Therefore, the comparison should not be interpreted as a claim of universal superiority over methods with richer inputs. Under this setting, ViCross achieves strong performance on JAAD\_beh and JAAD\_all. However, on PIE, methods equipped with richer structured cues can still perform better. For example, Faster-PCPnet~\cite{yang2024faster} achieves 0.94 Acc, compared with 0.82 Acc for ViCross.
These results indicate that external motion and structured perception cues remain beneficial in some scenarios, while ViCross provides a compact alternative that reduces frame-level perception requirements while maintaining competitive performance.
As shown in \figref{cloudrain}, ViCross consistently achieves strong performance across Acc, F1, Precision, and Recall over twelve random seeds. The colored scatter points and half-violin density estimates show that its results remain concentrated with limited variation, while the mean $\pm$ standard deviation markers further indicate stable performance. The Delta confidence intervals in the right panel also demonstrate that the improvements over the reproduced baselines are generally reliable. These results suggest that the gains of ViCross are robust rather than being caused by favorable random initialization. These multi-seed runs are separate from the seed-42 main tables.

\begin{figure}
\centering
\includegraphics[width=\textwidth]{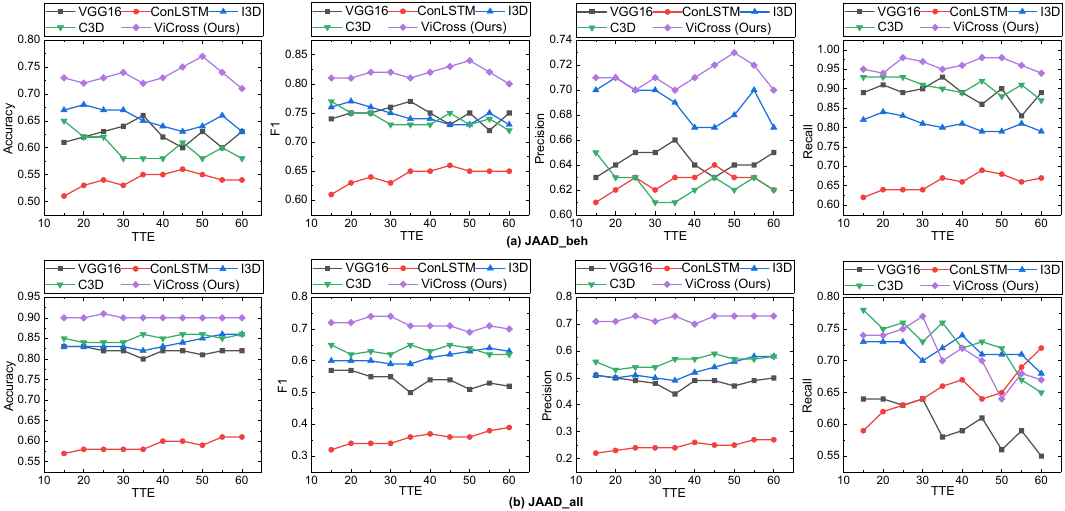}
\caption{Comparison with vision-centric methods under different prediction windows (15--60 frames) on JAAD\_all and JAAD\_beh.}
\label{TTE}
\end{figure}

\figref{TTE} further evaluates performance under different prediction horizons. As TTE increases, the task becomes more uncertain because pedestrian motion and scene context become harder to forecast. Most vision-centric baselines show degraded or fluctuating performance at longer horizons. In comparison, ViCross achieves higher accuracy, F1, and recall than competing vision-centric methods under most TTE settings on both JAAD\_beh and JAAD\_all.
\subsection{Ablation Study}

\subsubsection{Balancing Context Granularity and Target Resolution in Visual Tokenization}
\tableref{PSSIZE} analyzes the effect of VRPM design choices on prediction performance and efficiency. Enlarging the pedestrian crop from $112\times112$ to $224\times224$ decreases Acc from 0.73 to 0.69, suggesting that more local detail may introduce redundancy rather than improve target-centric reasoning. In contrast, increasing the panoramic resolution brings a slight performance gain, highlighting the importance of broader scene context. Patch granularity also affects the balance between prediction performance and computational cost: using coarser patches reduces inference time from 0.38s to 0.28s but weakens prediction performance. Overall, ViCross achieves a practical balance between accuracy and efficiency by preserving panoramic context, using a moderate pedestrian crop, and maintaining sufficient patch granularity. This result aligns with the motivation of VRPM, which allocates visual capacity between pedestrian details and scene context instead of uniformly increasing visual tokens.

\begin{table*}[t]
  \centering
  \caption{Design space exploration of the VRPM module. Different panoramic and pedestrian resolutions and coarse–fine patch sizes are evaluated in terms of trainable parameters (M) and their proportion (\%), inference time (s), and crossing prediction performance.}
  \label{PSSIZE}

  \resizebox{\textwidth}{!}{
    \begin{tabular}{cccccccccccc}
      \toprule
      Pan. Size & $P_{c}$ & Ped. Size & $P_{f}$ &
      Vis. Tokens & Trainable Para. & Ratio (\%) & Inference Time(s) &
      Acc & F1 & Prec & Rec \\
      \midrule
      224$\times$112 & 14 & 112$\times$112 & 7  & 32$\times$16  & 20.59 & 0.25\% &0.38  & 0.73 & 0.82 & 0.70 & 0.98 \\
      224$\times$112 & 14 & 224$\times$224 & 7  & 32$\times$16  & 20.59 & 0.25\% &0.46  & 0.69 & 0.79 & 0.69 & 0.91 \\
      448$\times$224 & 14 & 112$\times$112 & 7  & 128$\times$16 & 20.59 & 0.25\% &1.37  & 0.74 & 0.82 & 0.72 & 0.97 \\
      224$\times$112 & 28 & 112$\times$112 & 7  & 8$\times$16   & 26.61 & 0.32\% &0.28  & 0.66 & 0.78 & 0.65 & 0.97 \\
      224$\times$112 & 28 & 112$\times$112 & 14 & 8$\times$16   & 27.74 & 0.33\% &0.29  & 0.65 & 0.77 & 0.66 & 0.93 \\
      224$\times$112 & 28 & 224$\times$224 & 7  & 8$\times$16   & 26.61 & 0.32\% &0.33  & 0.64 & 0.76 & 0.65 & 0.92 \\
      \bottomrule
    \end{tabular}
  }
\end{table*}

\subsubsection{Influence of Auxiliary Objectives on Model Reliability}

\begin{figure}
  \centering
  \includegraphics[width=\textwidth]{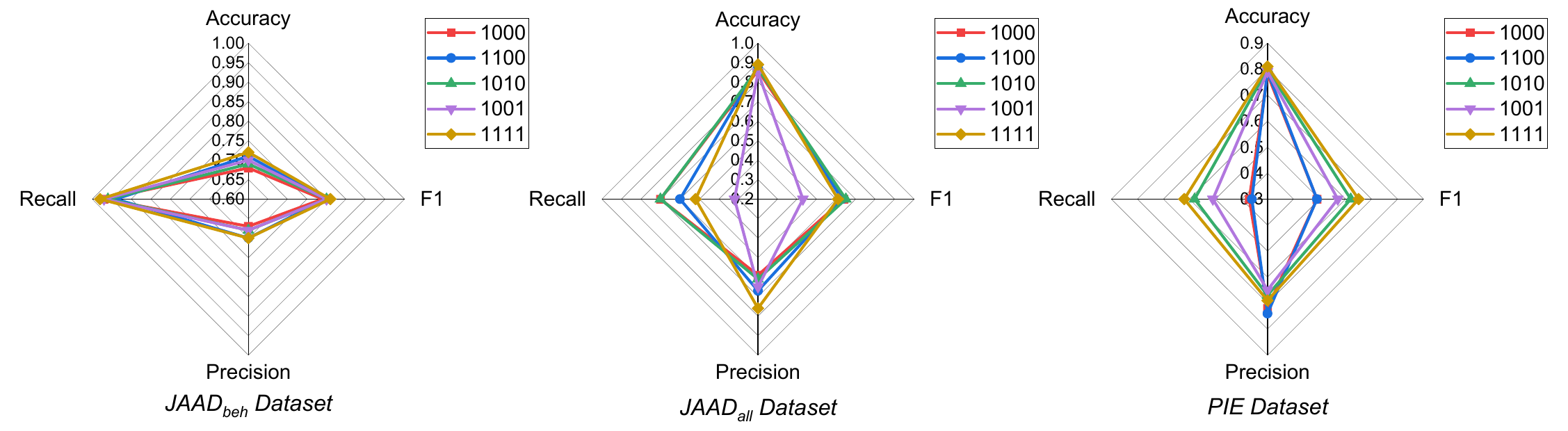}
  \caption{Performance comparison across three datasets with different loss weight combinations. Each plot shows results for one dataset, with the lines representing five loss weight configurations. The legend numbers (1000, 1100, 1010, 1001, 1111) correspond to the presence (1) or absence (0) of the generation ($\omega_1$), classification ($\omega_2$), prediction ($\omega_3$), and observation ($\omega_4$) losses, in that order.}
  \label{radar_loss}
\end{figure}

\figref{radar_loss} shows the effect of different auxiliary objective combinations on three benchmarks. Training with only the generation loss (1000) yields the weakest results, whereas enabling all objectives (1111) gives the best overall performance, reaching about 0.90 Acc on JAAD\_all and showing consistent gains on the other datasets. The intermediate settings (1100, 1010, 1001) generally fall between these two cases, indicating that each auxiliary objective contributes useful supervision rather than acting as a redundant loss term.
This suggests that the auxiliary objectives provide complementary supervision signals. Since these objectives are optimized on the shared representation before final generation, removing any of them weakens the spatial or semantic cues available for the final prediction.

\begin{figure}
\centering
\includegraphics[width=\textwidth]{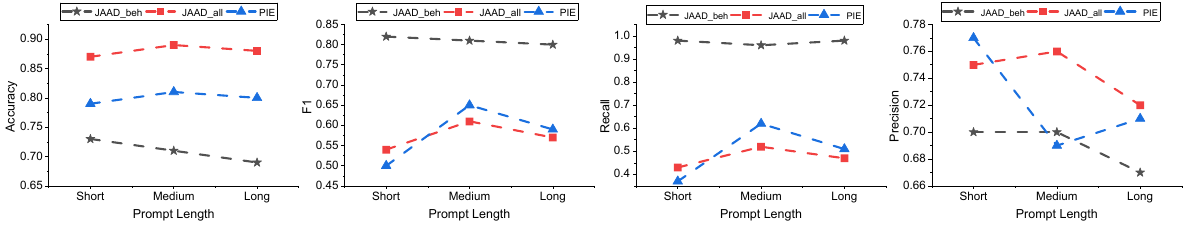}
\caption{Effect of different text prompt lengths on model performance across three datasets. Each subplot corresponds to one evaluation metric, illustrating how prompt length influences model performance.}
\label{prompt_zutu}
\end{figure}

\subsubsection{Effect of Prompt Complexity on Crossing Prediction}
\begin{figure}
\centering
\includegraphics[width=\textwidth]{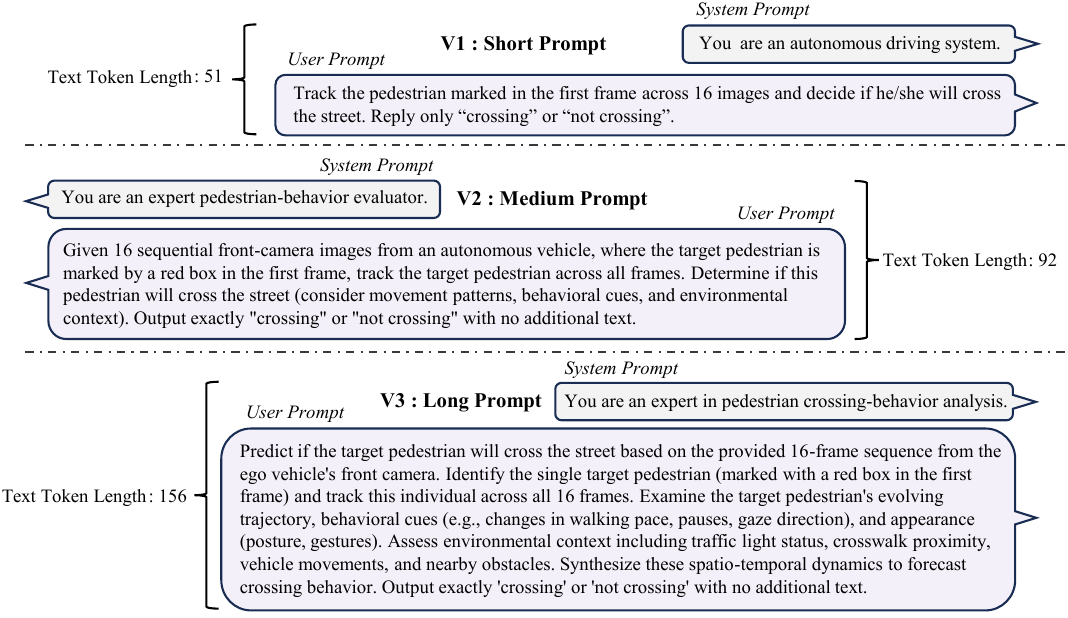}
\caption{Three prompt variants with progressively enriched semantic instructions are designed to investigate the impact of prompt complexity on vision-centric pedestrian crossing action prediction.}
\label{prompt}
\end{figure}

The results in \figref{prompt_zutu} indicate that prompt complexity affects performance, but the trend is not strictly linear across datasets. Overall, the medium prompt achieves a relatively favorable balance across the four metrics on all three datasets. As illustrated in \figref{prompt}, this prompt setting incorporates informative behavioral and contextual cues while remaining concise, enabling more stable reasoning about pedestrian motion. In contrast, further increasing prompt length does not lead to consistent gains. On JAAD\_beh, F1 and recall remain largely unchanged, while accuracy drops noticeably as prompt complexity increases. On PIE, both F1 score and recall decline when moving from the medium to the long prompt, and a similar degradation is observed on JAAD\_all. Overall, these results suggest that effective prompts should provide sufficient guidance without exceeding what can be reliably supported by visual evidence, with the medium prompt offering the most favorable balance.

\subsubsection{Scaling Effects of MLLM Backbones}
\begin{table}[ht]
\centering
\caption{Comparative analysis of different MLLM sizes in terms of performance, inference time, and parameter efficiency, where M and B denote million and billion parameters, respectively}
\resizebox{\textwidth}{!}{
    \begin{tabular}{ccccccccc}
    \toprule
    MLLM & Acc & F1 & Prec & Rec &
    Trainable Para. & Total & Ratio (\%) & Inference Time (s) \\
    \midrule
    Qwen2-VL-2B  & 0.72 & 0.80 & 0.71 & 0.92 & 11.13M & 2B  & 0.5\%  & 1.09 \\
    Qwen2-VL-7B  & 0.74 & 0.82 & 0.72 & 0.97 & 20.59M & 7B  & 0.25\% & 1.37 \\
    Qwen2-VL-72B & 0.72 & 0.81 & 0.70 & 0.98 & 107.22M & 72B & 0.15\% & 6.57 \\
    \bottomrule
    \end{tabular}
}
\label{MLLM}
\end{table}
Across MLLM scales, ~\tableref{MLLM} shows that performance does not increase monotonically with model size. The 7B model achieves the best overall result with 0.74 Acc and 0.82 F1, indicating that moderate scaling improves visual-textual reasoning for pedestrian crossing prediction. However, further scaling to 72B brings limited additional benefit while increasing inference time to 6.57 seconds. This suggests that the task is not simply limited by model capacity, but also by ambiguous visual evidence and temporal decision uncertainty. Therefore, the 7B model offers a more practical balance between prediction performance and computational cost.

\subsubsection{Fine-grained Scenario Evaluation}
\begin{figure}
\centering
\includegraphics[width=\textwidth]{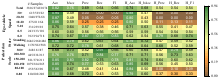}
\caption{Heatmap of ten evaluation metrics for ViCross across fine-grained PIE scenarios grouped by ego-vehicle speed, pedestrian state, and pedestrian scale. Scale is defined by the average pedestrian bbox height over the observation window. \# Samples denotes total/negative/positive samples.
}
\label{scenior}
\end{figure}


\begin{table*}[t!]
\begingroup
\centering
\caption{Quantitative breakdown under challenging conditions on JAAD\_beh. Small pedestrian denotes first-frame bbox height $<$ 100 px; occlusion denotes first-frame occlusion; crowded scene denotes windows with at least 10 pedestrians; non-clear weather denotes cloudy, rainy, or snowy conditions.}
\label{tiaozhan_senior}
\footnotesize
\renewcommand{\arraystretch}{0.9}
\begin{tabular*}{\textwidth}{@{\extracolsep{\fill}}llcccc@{}}
\toprule
Condition & Samples(total/negative/positive) & Acc & F1 & Prec & Rec \\
\midrule
Small pedestrian & 336/126/210 & 0.68 & 0.79 & 0.67 & 0.97 \\
Occlusion & 313/134/179 & 0.62 & 0.75 & 0.61 & 0.97 \\
Crowded scene & 545/209/336 & 0.66 & 0.78 & 0.65 & 1.00 \\
Non-clear weather & 1133/396/737 & 0.71 & 0.81 & 0.70 & 0.97 \\
\bottomrule
\end{tabular*}
\arrayrulecolor{black}
\endgroup
\end{table*}

As shown in \figref{scenior}, we evaluate ViCross on PIE under detailed scenario conditions by stratifying the test set according to ego speed, pedestrian state, and pedestrian scale. The figure reports standard metrics, balanced accuracy, and the Hard protocol~\cite{rasouli2024diving}, which requires correct and consistent decisions across all prediction windows. ViCross achieves 0.82 Acc but 0.74 H\_Acc on PIE, indicating that temporally consistent prediction remains harder than single-window prediction. The main drops occur under high ego speed and small pedestrian scale. At high speeds, limited positive samples make Acc less reliable, while small pedestrians weaken orientation, motion, and identity cues, reducing Acc to 0.80.
\tableref{tiaozhan_senior} further evaluates ViCross on JAAD\_beh under challenging scenarios, including small pedestrians, occlusion, crowded scenes, and non-clear weather. The results show performance drops in these cases, suggesting that limited visibility and scene ambiguity make crossing prediction more difficult. The high Rec but lower Prec further indicates that ViCross tends to preserve sensitivity to crossing cases, at the cost of more false positives under challenging conditions.
\tableref{hard_suiji} reports small deviations and narrow 95\% confidence intervals across twelve seeds, confirming that the Hard-protocol degradation is stable across initializations.

\begin{table}[htbp]
\centering
\caption{Hard-protocol reliability analysis on PIE under two representative settings over twelve random seeds. Samples are reported as total/negative/positive, results as mean $\pm$ standard deviation. $\Delta$Acc measures the drop from standard Acc to H\_Acc, with CI denoting its 95\% confidence interval.}


\resizebox{\textwidth}{!}{
\begin{tabular}{lccc cccc c}
\toprule
\multirow{2}{*}{Setting}
& \multicolumn{3}{c}{Samples}
& \multirow{2}{*}{H\_Acc}
& \multirow{2}{*}{H\_Prec}
& \multirow{2}{*}{H\_Rec}
& \multirow{2}{*}{H\_F1}
& \multirow{2}{*}{$\Delta$Acc 95\% CI} \\
\cmidrule(lr){2-4}
& total & negative & positive
& & & & & \\
\midrule

scale 0--80
& 810 & 602 & 208
& 0.82 $\pm$ 0.06
& 0.55 $\pm$ 0.21
& 0.40 $\pm$ 0.12
& 0.42 $\pm$ 0.06
& [-0.01, 0.04] \\

state walking
& 1170 & 582 & 588
& 0.61 $\pm$ 0.05
& 0.67 $\pm$ 0.07
& 0.42 $\pm$ 0.15
& 0.50 $\pm$ 0.12
& [0.20, 0.24] \\

\bottomrule
\end{tabular}
}

\arrayrulecolor{black}

\label{hard_suiji}
\end{table}

\subsubsection{Generalization Capability Across Diverse Urban Scenarios}

\begin{table}[t]
\centering
\caption{Cross-dataset generalization results with training on JAAD\_all and evaluation on JAAD\_all, JAAD\_beh, and PIE. Performance is reported for the proposed method and other vision-centric baselines.}
\label{cross-dataset}

\footnotesize
\setlength{\tabcolsep}{3pt}

\begin{tabular*}{\columnwidth}{@{\extracolsep{\fill}}lcccccccccccc@{}}
\toprule
& \multicolumn{12}{c}{JAAD\_all} \\
\cmidrule(lr){2-13}

& \multicolumn{4}{c}{JAAD\_all}
& \multicolumn{4}{c}{JAAD\_beh}
& \multicolumn{4}{c}{PIE} \\

\cmidrule(lr){2-5}
\cmidrule(lr){6-9}
\cmidrule(lr){10-13}

Method
& Acc & F1 & Prec & Rec
& Acc & F1 & Prec & Rec
& Acc & F1 & Prec & Rec \\
\midrule

VGG16   & 0.82 & 0.55 & 0.49 & 0.63 & 0.57 & 0.63 & 0.68 & 0.58 & 0.62 & 0.24 & 0.28 & 0.22 \\
ConLSTM & 0.63 & 0.32 & 0.24 & 0.48 & 0.54 & 0.64 & 0.62 & 0.67 & 0.63 & 0.27 & 0.31 & 0.24 \\
I3D     & 0.81 & 0.63 & 0.66 & 0.61 & 0.65 & 0.72 & 0.73 & \textbf{0.71} & 0.64 & \textbf{0.36} & \textbf{0.35} & \textbf{0.36} \\
C3D     & 0.84 & 0.65 & 0.57 & \textbf{0.75} & 0.66 & 0.72 & 0.74 & \textbf{0.71} & 0.61 & 0.27 & 0.28 & 0.25 \\
ViCross & \textbf{0.90} & \textbf{0.71} & \textbf{0.73} & 0.69
        & \textbf{0.74} & \textbf{0.77} & \textbf{0.86} & 0.69
        & \textbf{0.67} & 0.15 & 0.28 & 0.11 \\

\bottomrule
\end{tabular*}
\end{table}

To evaluate the transferability of ViCross across datasets, we conduct a cross-dataset study. As shown in \tableref{cross-dataset}, ViCross performs well on JAAD\_all and maintains competitive results on JAAD\_beh, suggesting reasonable transferability under a moderate distribution shift.
When transferred to PIE, ViCross still retains moderate accuracy but shows a clear drop in recall and F1 score. This indicates that the model becomes less sensitive to positive crossing cases under a stronger domain shift, rather than demonstrating robust generalization in all scenarios. The degradation may be related to differences in scene distribution, pedestrian scale, motion patterns, and recording conditions between JAAD and PIE. These results suggest that ViCross can capture transferable target-centric cues, but its robustness under severe cross-dataset shifts remains a limitation to be further improved.

\begin{table}[t]
\centering
\caption{Ablation study of VRPM and SCES on JAAD\_{beh}.}

\small
\renewcommand{\arraystretch}{0.9}

\begin{tabular*}{\columnwidth}{@{\extracolsep{\fill}}cccccc@{}}
\toprule
VRPM & SCES & Acc & F1 & Prec & Rec \\
\midrule
$\times$     & $\times$     & 0.71 & 0.79 & 0.71 & 0.90 \\
$\checkmark$ & $\times$     & 0.72 & 0.80 & 0.72 & 0.90 \\
$\times$     & $\checkmark$ & 0.72 & 0.81 & 0.70 & 0.96 \\
$\checkmark$ & $\checkmark$ & \textbf{0.74} & \textbf{0.82} & \textbf{0.72} & \textbf{0.97} \\
\bottomrule
\end{tabular*}

\label{ablation}
\end{table}

\subsubsection{Analyzing the Impact of Key Design Components}

\tableref{ablation} provides component-level evidence for the two proposed modules. 
When VRPM is enabled alone, accuracy increases from 0.71 to 0.72 and F1 from 0.79 to 0.80. This shows that adaptive target-centric tokenization improves the spatial representation of the target pedestrian, although the overall prediction tendency changes only moderately. 
When SCES is enabled alone, recall increases from 0.90 to 0.96 while accuracy remains stable at 0.72, indicating that spatially guided training makes the model more sensitive to crossing-related cues. Combining VRPM and SCES achieves the best overall performance, with 0.74 Acc, 0.82 F1, and 0.97 Rec. These results show that VRPM and SCES contribute through different but complementary mechanisms: VRPM strengthens the target-centric visual representation, while SCES further guides the shared representation with spatial supervision during training.

\begin{table*}[htp]
\centering
\caption{Internal ablation of the feature fusion strategy in VRPM on the JAAD\_beh dataset. All variants use the same panoramic resolution, pedestrian-region resolution, and patch sizes.}
\label{tab:vrpm_fusion_ablation}

\begingroup

\footnotesize
\renewcommand{\arraystretch}{0.9}

\begin{tabular*}{\columnwidth}{@{\extracolsep{\fill}}lcccc@{}}
\toprule
Fusion Strategy & Acc & F1 & Prec & Rec \\
\midrule
Coarse panoramic feature only & 0.69 & 0.80 & 0.70 & 0.92 \\
Fine feature replacement & 0.70 & 0.81 & 0.68 & \textbf{0.99} \\
Direct addition & 0.72 & 0.81 & 0.69 & 0.97 \\
Adaptive fusion (ours) & \textbf{0.74} & \textbf{0.82} & \textbf{0.72} & 0.97 \\
\bottomrule
\end{tabular*}

\arrayrulecolor{black}
\endgroup

\end{table*}

\subsubsection{Internal Ablation Study on VRPM Feature Fusion Strategy}
To verify the proposed content adaptive fusion, we compare four variants, including coarse panoramic features only, pooled fine grained features, direct addition of coarse and fine features, and adaptive fusion. As shown in Table~\ref{tab:vrpm_fusion_ablation}, coarse features alone perform worst due to insufficient pedestrian details. Fine grained features improve target representation, and direct addition further combines local and contextual cues but may introduce redundant responses. In contrast, adaptive fusion achieves the best performance on most metrics by selectively injecting fine-grained cues into panoramic tokens, demonstrating the effectiveness of adaptive feature integration in VRPM.

\subsubsection{Temporal-Density Ablation of Observation Localization in SCES}
We conduct a temporal density ablation on the observation localization constraint to further analyze SCES. Different from the objective ablation in \figref{radar_loss}, this experiment keeps all SCES objectives unchanged and only varies the observed frames supervised by $L_{OBS}$. As shown in Table~\ref{tab:sces_temporal_density}, supervising only the last frame provides a limited spatial anchor, while sparse frame supervision adds intermediate guidance but still leaves part of the sequence weakly constrained. Supervising all observed frames achieves the best overall performance, showing that denser spatial guidance better preserves target consistency over time. These results support SCES as a training strategy for proactive spatial rectification of shared visual representations.

\begin{table}[t]
\centering
\begingroup
\caption{
Internal ablation of SCES on the temporal density of observation localization on JAAD\_beh. All SCES objectives are kept unchanged, and only the frames supervised by \(L_{OBS}\) are varied.}
\label{tab:sces_temporal_density}

\footnotesize
\renewcommand{\arraystretch}{0.9}

\begin{tabular*}{\columnwidth}{@{\extracolsep{\fill}}p{0.34\columnwidth}p{0.22\columnwidth}cccc@{}}
\toprule
Observation Localization & Supervised Frames & Acc & F1 & Prec & Rec \\
\midrule
Last-frame bbox
& \(\{T\}\)
& 0.71 & 0.81 & 0.71 & 0.94 \\

Sparse-frame bbox
& \(\{3,7,10,13,\ldots,T\}\)
& 0.72 & 0.80 & 0.71 & 0.91 \\

Full-frame bbox (ours)
& \(\{1,2,\ldots,T\}\)
& \textbf{0.74} & \textbf{0.82} & \textbf{0.72} & \textbf{0.97} \\
\bottomrule
\end{tabular*}

\endgroup
\end{table}

\subsubsection{Generator-based versus Classifier-based Final Decision}

\begin{table*}[t]
\centering
\begingroup

\caption{Comparison of classifier and generator decisions. The same trained ViCross model is used, with the final decision taken from either the auxiliary classifier or the MLLM generated output.}
\label{tab:generator_classifier_ablation}

\resizebox{\textwidth}{!}{%
\begin{tabular}{lcccccccccccc}
\toprule
\multirow{2}{*}{Final Decision} 
& \multicolumn{4}{c}{JAAD\_beh} 
& \multicolumn{4}{c}{JAAD\_all} 
& \multicolumn{4}{c}{PIE} \\
\cmidrule(lr){2-5} \cmidrule(lr){6-9} \cmidrule(lr){10-13}
& Acc & F1 & Prec & Rec 
& Acc & F1 & Prec & Rec 
& Acc & F1 & Prec & Rec \\
\midrule
Classifier-based 
& 0.70 & 0.79 & 0.69 & 0.91 
& 0.72 & 0.51 & 0.37 & \textbf{0.81} 
& 0.78 & 0.62 & 0.66 & 0.58 \\

Generator-based 
& \textbf{0.74} & \textbf{0.82} & \textbf{0.72} & \textbf{0.97} 
& \textbf{0.90} & \textbf{0.71} & \textbf{0.73} & 0.69 
& \textbf{0.82} & \textbf{0.66} & \textbf{0.69} & \textbf{0.64} \\
\bottomrule
\end{tabular}%
}
\endgroup
\end{table*}

\begin{figure}
\centering
\includegraphics[width=\textwidth]{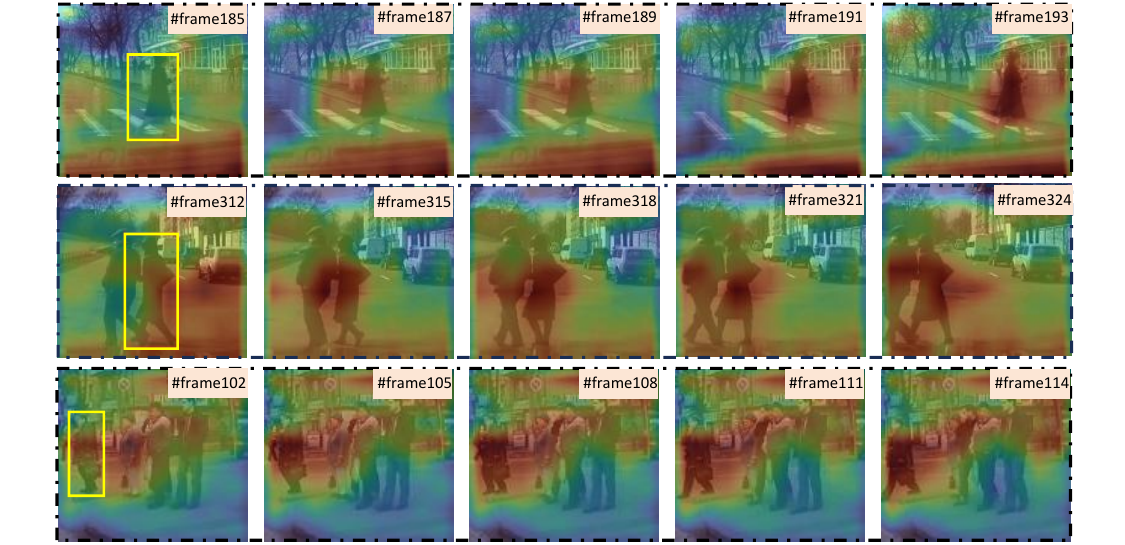}
\caption{Visualization of weighted graph from the content aware stage of the VRPM. Warmer colors mark regions the model deems important and therefore allocates finer-grained pedestrian information to enhance, while cooler colors indicate less important regions that are mainly represented by coarse-grained features across frames.}
\label{heat}
\end{figure}

\begin{figure*}[htbp]
\centering
\includegraphics[width=\textwidth]{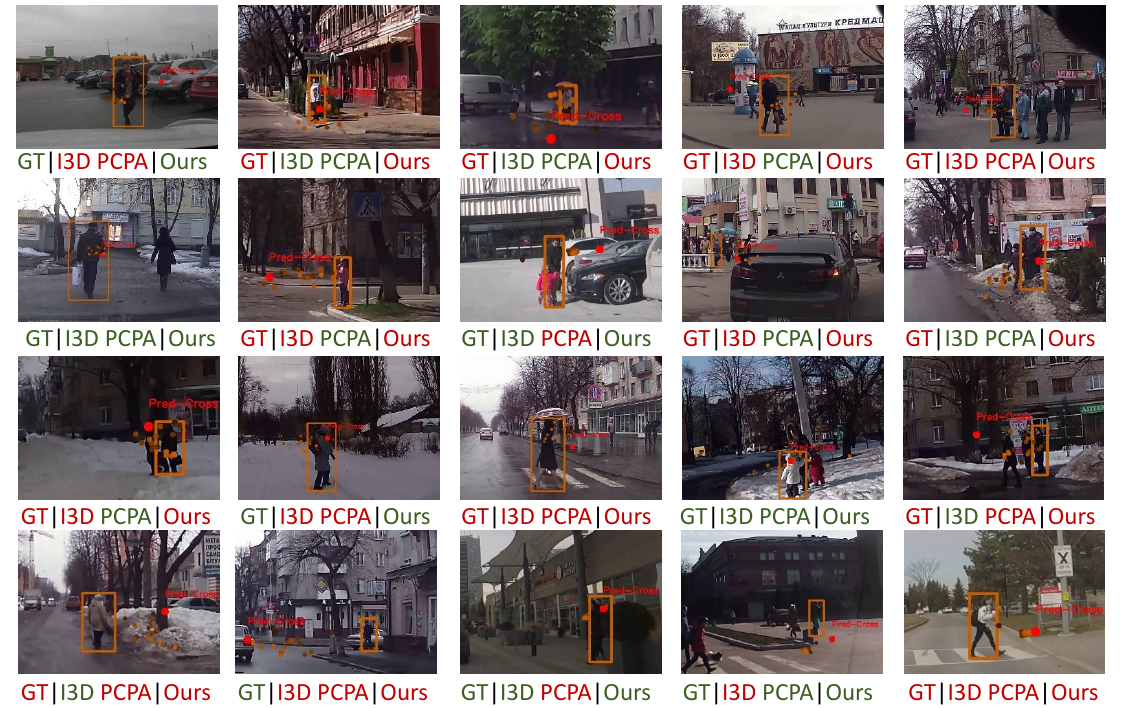}
\caption{Qualitative visualization of pedestrian crossing action prediction. GT denotes the ground-truth label. We compare I3D and PCPA against ViCross. \textcolor{red!60!black}{Red} indicates crossing and \textcolor{green!50!black}{green} indicates non-crossing, a method is considered correct when its color matches the GT. \textcolor{orange!70!black}{The orange dots} show ViCross’s tracked pedestrian positions over the past 16 frames, where lighter colors correspond to more recent frames. The \textcolor{red!60!black}{red dot} marks ViCross’s predicted event occurrence location.}
\label{big_vis}
\end{figure*}

To justify text generation as the final decision mechanism, we compare two inference variants of the same trained ViCross model: one uses the MLLM-generated textual output, and the other uses the auxiliary binary classifier. As shown in Table~\ref{tab:generator_classifier_ablation}, the generator-based decision outperforms the classifier-based counterpart on most metrics, demonstrating its overall advantage. This indicates that the generation branch provides a more effective prompt-conditioned decision path by jointly leveraging visual evidence, task instructions, and learned semantic constraints. In contrast, the classifier mainly acts as an auxiliary discriminative regularizer for shared visual tokens, and using it as the final head degrades performance.

\subsubsection{Qualitative Analysis and Model Interpretation}

\begin{figure}
\centering
\includegraphics[width=\textwidth]{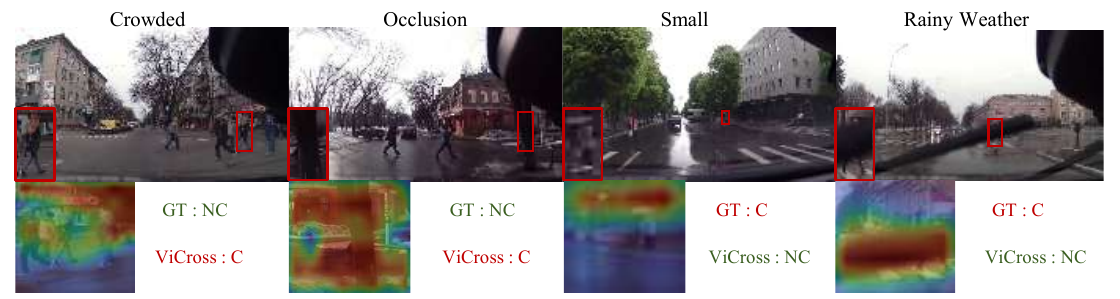}
\caption{Representative failure cases of ViCross under crowded, occluded, small-scale, and adverse weather conditions. Red boxes denote target pedestrians. The attention maps show that the model fails to maintain target-centric focus, leading to incorrect predictions.}
\label{failure}
\end{figure}

The energy map in \figref{heat} shows how VRPM adaptively allocates fine-resolution capacity during coarse-fine feature fusion. Warmer regions indicate stronger contributions from fine-grained pedestrian-region features, while cooler regions rely more on coarse panoramic context. High responses mainly appear around the target pedestrian and crossing-relevant regions, such as the curbside and walking corridor, whereas static backgrounds receive weaker responses. This suggests that VRPM selectively injects fine-grained information into decision-relevant regions, supporting target-centric reasoning without densely refining the entire frame. The smooth response changes further indicate temporally consistent allocation along pedestrian motion.
\figref{big_vis} compares ViCross with I3D and PCPA for pedestrian crossing action prediction. ViCross produces predictions more consistent with ground-truth action labels, with smoother trajectories and future event locations better aligned with the target pedestrian's near-future motion. These results indicate more stable spatial reasoning, while ViCross requires only first-frame target bounding box initialization, unlike I3D and PCPA, which rely on frame-level annotations.
\figref{failure} shows representative failure cases under challenging conditions, where errors mainly arise from weakened target-centric focus in crowded scenes, severe occlusion, small-target cases, and adverse weather.

\section{Limitations}

Despite its simplified design, ViCross still relies on first-frame target specification, making it sensitive to inaccurate initialization, severe occlusion, small or far-away pedestrians, and crowded scenes where neighboring pedestrians may interfere with target identification. Although SCES improves cross-frame spatial consistency, it does not explicitly correct localization during inference, so target drift may remain in crowded or ambiguous scenes. Beyond localization, the MLLM backbone introduces non-negligible inference cost compared with lightweight task-specific models, and transfer to PIE shows remaining sensitivity to severe domain shifts, especially in recall and F1. Future work will explore lightweight acceleration, broader training data, and domain adaptation to improve efficiency and robustness.

\section{Conclusion}

We present ViCross, a vision-centric pedestrian crossing action prediction framework powered by MLLMs, built upon Variable Resolution Patch Mapping and a Spatial Constraint Enhancement Strategy. By focusing visual encoding on the target pedestrian and enforcing cross-frame spatial consistency, ViCross enables target-centric action reasoning from video frames using only first-frame target initialization. Extensive experiments show that ViCross outperforms vision-centric baselines and remains competitive with multi-source approaches in several settings, despite requiring substantially fewer frame-level external perception cues.
Specifically, ViCross initializes the target with a single bounding box in the first frame, simplifying supervision and avoiding detector- or tracker-based pipelines during inference. While this setting prevents the framework from being strictly end-to-end, future work will investigate automatic target initialization to further reduce external dependencies.

\section*{Author Contributions}
Yao Tian: Writing – original draft, data curation, methodology, software.  
Le Yang: Conceptualization, writing – review \& editing, methodology, funding acquisition, supervision.  
Binglu Wang: Writing – review \& editing, funding acquisition, resources.

\section*{Declaration of generative AI and AI-assisted technologies in the manuscript preparation process.}
During the preparation of this work, the authors used GPT 5.5 in order to improve language and readability. After using this tool, the authors review and edit the content as needed and take full responsibility for the content of the published article.

\section*{Acknowledgments}
This work is supported by the National Natural Science Foundation of China [grant numbers 62401447 and 62306101]; the Fundamental Research Funds for the Central Universities CHD [grant number 300102325102]; the Key Research and Development Program of Shanxi [grant number 2024CY2-GJHX-08]; and the State Key Laboratory of Respiratory Health and Multimorbidity [grant number SKLRHM01213].

\bibliographystyle{elsarticle-num}

\bibliography{cas-refs-PR_suo}

\end{document}